\documentclass[conference]{IEEEtran}
\IEEEoverridecommandlockouts

\usepackage{hyperref}       
\usepackage{url}            
\usepackage{booktabs}       
\usepackage{amsfonts}       
\usepackage{nicefrac}       
\usepackage{microtype}      
\usepackage{xcolor}         
\usepackage{graphicx}       
\usepackage{tabularx}
\usepackage{algorithm}      
\usepackage[noend]{algpseudocode}

\usepackage{amsmath}
\usepackage{amsthm}
\usepackage{lipsum}
\usepackage[numbers]{natbib}
\usepackage{setspace}
\usepackage{titlesec}
\titlespacing*{\section}{0pt}{*0.5}{*0.5}
\titlespacing*{\subsection}{0pt}{*0.5}{*0.5}

\title{\LARGE Generative Expansion of Small Datasets: An Expansive Graph Approach}

\usepackage{titlesec}

\makeatletter
\newcommand{\linebreakand}{%
  \end{@IEEEauthorhalign}
  \hfill\mbox{}\par
  \mbox{}\hfill\begin{@IEEEauthorhalign}
}
\makeatother

\author{
  \begin{tabular}{@{}c@{\hspace{1em}}c@{\hspace{1em}}c@{\hspace{1em}}c@{}}
    Vahid Jebraeeli & Bo Jiang & Hamid Krim & Derya Cansever \\
    ECE Department & Neuroscience Department & ECE Department & ECE Department \\
    NC State University & Washington University & NC State University & NC State University \\
    Raleigh, USA & St. Louis, USA & Raleigh, USA & Raleigh, USA \\
    vjebrae@ncsu.edu & bo.j@wustl.edu & ahk@ncsu.edu & dhcansev@ncsu.edu
  \end{tabular}
}
\begin{document}
\maketitle
\begin{abstract}
Limited data availability in machine learning significantly impacts performance and generalization. Traditional augmentation methods enhance moderately sufficient datasets. GANs struggle with convergence when generating diverse samples. Diffusion models, while effective, have high computational costs. We introduce an Expansive Synthesis model generating large-scale, information-rich datasets from minimal samples. It uses expander graph mappings and feature interpolation to preserve data distribution and feature relationships. The model leverages neural networks' non-linear latent space, captured by a Koopman operator, to create a linear feature space for dataset expansion. An autoencoder with self-attention layers and optimal transport refines distributional consistency. We validate by comparing classifiers trained on generated data to those trained on original datasets. Results show comparable performance, demonstrating the model's potential to augment training data effectively. This work advances data generation, addressing scarcity in machine learning applications. \footnote{Thanks to the generous support of ARO grant W911NF-23-2-0041.}

\end{abstract}


\section{Introduction}

Data scarcity in machine learning limits performance and generalizability across various fields. Current augmentation techniques and GANs lack diversity and stability. Our "Expansive Synthesis" addresses this shortfall by providing a framework to create extensive datasets from limited data. This approach improves data sufficiency for effective training. It builds on "dimension expansion" \citep{DBLP:journals/corr/ForbesG14}, "expander graphs" \citep{margulis1973explicit}, self-attention \citep{vaswani2017attention}, and optimal transport \citep{villani2009optimal} for well-grounded synthesis.
Data augmentation in deep learning uses geometric and color transformations to enhance the generalization capabilities of CNNs. Random Oversampling (ROS) and Synthetic Minority Over-sampling Technique (SMOTE) \citet{chawla2002smote} address class imbalances but have limitations. GANs \citet{goodfellow2014generative} produce high-quality synthetic data, with improvements from DCGANs \citet{radford2015unsupervised} and CycleGANs \citet{zhu2017unpaired}. Neural Style Transfer \citet{gatys2015neural} and Fast Style Transfer \citep{johnson2016perceptual} add artistic dimensions. However, GANs face instability and diversity issues. Meta-learning approaches like Neural Architecture Search (NAS) and AutoAugment \citep{cubuk2019autoaugment} automate augmentation strategies but are complex and better suited for smaller-scale tasks.

Motivated by data-starved environments encountered in practice, we seek a Neural Network's capacity to extrapolate from latent space representations of minimal data. Towards lifting the limited data augmentation strategies for severely small datasets,  we propose a dimension expansion-inspired synthesis we refer to as "Expansive Synthesis" to generate large-scale, highly training-efficient datasets from small samples. Utilizing regularization techniques like self-attention \citep{vaswani2017attention}, our approach leverages dimension expansion and its expander graph perspective \citep{DBLP:journals/corr/ForbesG14} to exploit features for robust data synthesis, facilitating model training.
This approach is in some sense the dual of our   recently data condensation \citep{jebraeeli2024koopcon}. We describe a framework where extracted and attention-driven features define a Koopman space, and optimally combine nonlinear components for dimension expansion while preserving data distribution via optimal transport. Section 2 contextualizes the deep learning of a Koopman Operator\citep{koopman1931hamiltonian, dey2023dlkoopman}. Section 3 details the Expansive Synthesis model. Section 4 presents experimental validation, demonstrating the effectiveness in generating expansive datasets and on evaluating classifier performance. Section 5 concludes, summarizing findings and outlining future directions.

\section{Related Background}

A Koopman operator\cite{koopman1931hamiltonian} enables a linear analysis of nonlinear dynamical systems. We adopt a deep learning model estimate of a Koopman operator whose generated data distribution is preserved by an  Optimal Transport (OT) transformation.

\subsection{Koopman Feature Space}
Koopman theory offers an elegant and simplified framework for analyzing nonlinear dynamical systems in a linear space of functions of state variables\citep{koopman1931hamiltonian}. Building on this theory, a deep learning model was proposed for approximating a Koopman operator, thereby providing flexible application in data-driven problems\cite{dey2023dlkoopman}.  

\subsection{Data Distribution Preservation}
It is critical to preserve the distribution of a synthesized expanded dataset to safeguard its utility for further training (or other) purposes. To that end, Wasserstein distance is used for regularization, as illustrated in Fig.~\ref{fig:1}(b). Additional details will be discussed in the Methodology section. 

\section{Methodology}
\begin{figure}
  \centering
  \includegraphics[width=9cm]{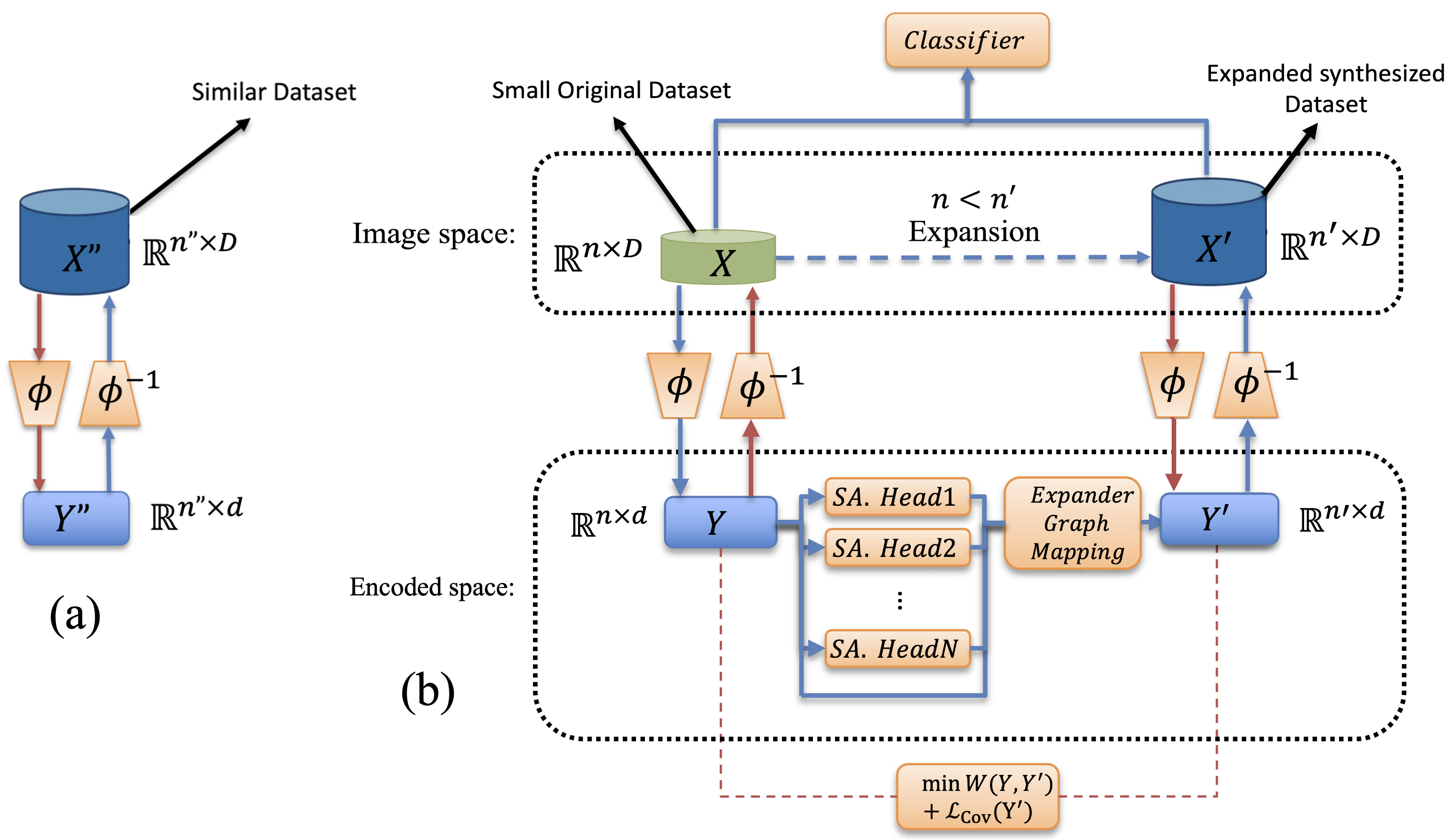}
  \vspace{-20pt}
  \caption{Overall architecture of Expansive Synthesis model. 
  \textbf{(a)} Pretraining phase using a similar larger dataset \(X''\) to learn general features, which are encoded and decoded to produce \(Y''\). 
  \textbf{
  (b)} Fine-tuning phase on the smaller minimal sample dataset \(X\) to adapt the model's weights, followed by the expansion of \(X\) to generate the synthesized dataset \(X'\) using expander graph mapping and self-attention mechanisms. The expanded dataset \(X'\) is then used to train a classifier.}
  \label{fig:1}
\end{figure}

Pursuing our objective primarily entails, \\
{\bf Claim:} {\bf \it A near-optimal and feature consistent expansion \(X'\) of a dataset \(X\) can be achieved in a Koopman-data space using Expander Graph Mapping, with proper distribution and feature refinement regularizations.}\\
Fig.~\ref{fig:1} illustrates  our proposed method to expand a minimal dataset \(X\) into the a larger and feature-rich data set \(X'\), with key attributes preserved by a Koopman-representation linear evolution of  non-linear dynamics \cite{koopman1931hamiltonian}. We fist obtain a latent representation \(Y \in \mathbb{R}^{n \times d}\) via \(\phi:\mathbb{R}^{n \times D} \rightarrow \mathbb{R}^{n \times d}\) (\(d<D\)). Multi-head Spatial Self-Attention captures discriminative features, while Expander Graph Mapping \cite{dey2023dlkoopman} generates diverse datapoints, maintaining original distribution via Wasserstein distance and Covariance loss.
To pursue the dimension expansion of small but diverse data sets, we first define a functional space where a meaningful preservation of non-linear features and characteristic of training data at hand.
To that end, as illustrated on the left-hand side of Fig.~\ref{fig:1}(b), we aim to glean a Koopman-like characterization of data across all classes of interest as defined above. 
A training data sample as an aggregate of extracted features (possibly refined by an attention mechanism), may be interpreted as a structured set of graph nodes, benefiting of an expander graph diversification strategy to achieve a potential dimension expansion\citep{DBLP:journals/corr/ForbesG14} facilitating our proposed approach as discussed next.  

\begin{figure}
  \vspace{-10pt}
  \centering
  \includegraphics[width=9cm]{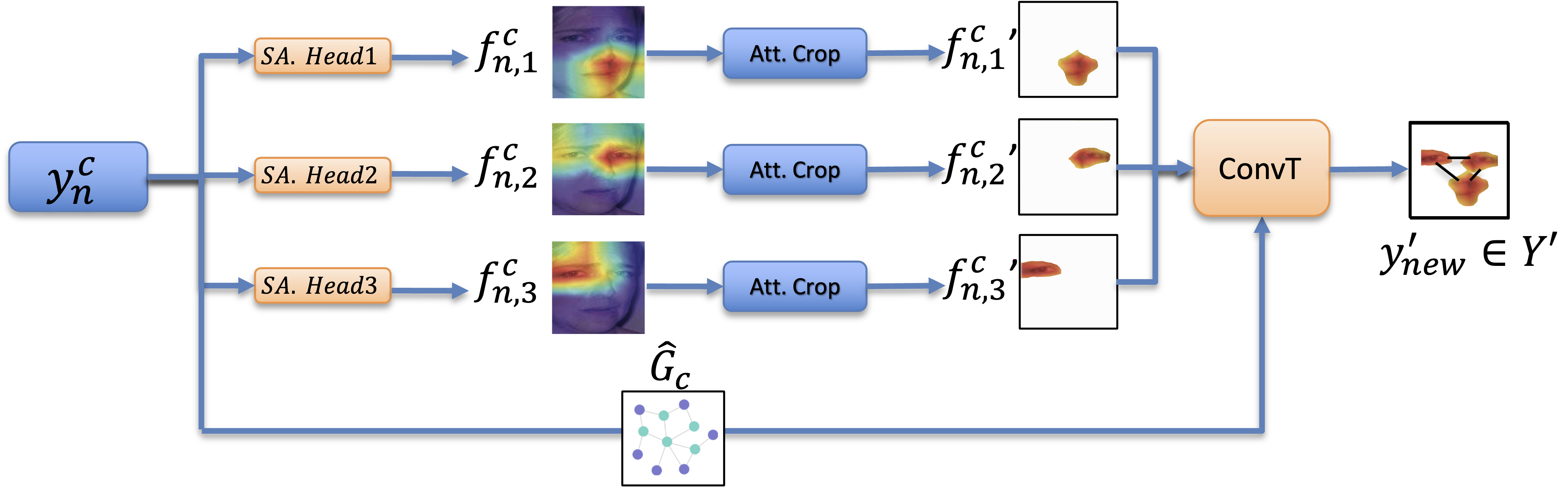}
  \vspace{-20pt}
  \caption{ Architecture of Expander Graph Mapping}
  \vspace{-20pt}
  \label{fig:2}
\end{figure}

\subsection{Autoencoder:Koopman Feature Space}
Our goal of a Data "Expansive Synthesis" is initiated by pre-training a convolutional Autoencoder (AE) uses a dataset \( X'' \) analogous to (and distinct from) the target smaller dataset \( X \). This phase equips the AE with understanding of general features in a typically sufficient dataset. The AE's architecture has multiple convolutional layers, each with a set of \(cout-number\) filters and associated parameters \( W_1, W_2, ..., W_{cout} \), optimized to capture and encode essential data characteristics \cite{goodfellow2014generative, krizhevsky2012imagenet}. With \( W_i \) as the \(i^{th}\) filter parameters, a corresponding  convolutional transformation  is defined as \( \text{conv}(X''; W_i) \). Intrinsically meant for  dimension reduction, the AE encodes \(X''\) to \( Y'' \), to retain key features as a compact and informative representation  by minimizing a reconstruction loss. As shown in Fig.~\ref{fig:1}(a), the encoding is carried out by a mapping  \( \phi \), with a corresponding decoding  using \( \phi^{-1} \).
In adapting the pre-trained AE to a Koopman-Autoencoder (KAE), we optimize the pre-trained weights \( W \) to capture $X$ data characteristics. Our model introduces scaling (\( \gamma \)) and shifting (\( \beta \)) parameters to fine-tune the KAE' performance, to yield the following convolution operation during fine-tuning,

\begin{minipage}{8.5cm}
\begin{equation}
\small
\begin{aligned}
&\text{conv}(X; W) \cdot \gamma + \beta = \\
&\text{conv} \big( X; \{ \gamma_1 W_1 + \beta_1, \ldots, \gamma_{cout} W_{cout} + \beta_{cout} \} \big).
\end{aligned}
\end{equation}
\end{minipage}

Here, \( \gamma_i \) adjusts each filter's activation strength, controlling input feature sensitivity \cite{he2016deep}. Higher \( \gamma_i \) enhances filter response to specific features; \( \beta_i \) modifies activation threshold. These filter-specific parameters enable precise adaptation to the smaller dataset \( X \). Fine-tuning maintains generalization from the larger similar dataset while adapting to \( X \)'s distinct features. This refined AE generates the expanded synthesized dataset \( X' \) (Fig.~\ref{fig:1}(b), right), transitioning from encoded representation \( Y' \) to image space via \( \phi^{-1} \).

\subsection{Feature Refinement: A Self-Attention Mechanism}
We address the incompatibility of  longitudinal learning of attention maps across  token sequences\cite{vaswani2017attention} with images. As shown in Fig.~\ref{fig:1}(b), our proposed spatial multi-head SA mechanism to improve the resulting capture of spatial dependencies, entails the  partitioning of encoded data into patches, and the proximal positioning of flattened features in a stacked representation (with no positional encoding).  This in fact enhances the spatial self-attention across all dimensions to focus on interacting features throughout a given image.

\subsection{Expander Graph Mapping}
We invoke the notion of Expander Graph Mapping\footnote{An expander graph is a well connected sparse graph, with every not-too large subset of vertices has a large boundary (hence with a capacity of diverse connections for new generation). } to pursue a dimension expansion and realize our Expansive Synthesis (ES) model. The ES follows a systematic selection of a proper feature set in the encoded space $Y$ which are structurally combined to generate a new and larger dataset \(\mathbf{Y}'\) with new datapoints as row vectors \({y'}_{\text{new}}\). \\
\textbf{Proposition:} {\bf \it A composition map of a learned attention transformation and of an Expander Graph Mapping achieve a dimension expansion of $Y\in {\mathbb R}^{n\times d}$ to $Y'\in {\mathbb R}^{n'\times d}$ where $n'>n$ is of  ${\mathcal O}(n^L)$, with $L$ as the uniform feature-size selected for each datapoint in $Y\in {\mathbb R}^{n\times d}$.}\\
In order to formalize this dimension expansion process, we define it as a composition of two maps. Let ${\mathcal A}: {\mathbb R}^{n\times d} \rightarrow {\mathbb R}^{n\times d\times L}$ and ${\mathcal E_G}: {\mathbb R}^{n\times d\times L} \rightarrow {\mathbb R}^{n'\times d}$ be two maps. The dimension expansion can be then expressed as:
\begin{minipage}{8.5cm}
\begin{equation}
    \begin{aligned}
    ({\mathcal E_G} \circ {\mathcal A})(Y) = {\mathcal E_G}({\mathcal A}(Y)) = Y' \in {\mathbb R}^{n'\times d}, \quad \forall Y \in {\mathbb R}^{n\times d}
    \end{aligned}
\end{equation}
\end{minipage}
The first map, ${\mathcal A}$, represents the application of self-attention to extract $L$ important features from each datapoint. The second map, ${\mathcal E_G}$, represents the Expander Graph Mapping that uses these features to generate new datapoints, resulting in the expanded dataset.
Each row vector  \(y_i\) in \(Y \in \mathbb{R}^{n \times d}, i=1,\cdots, n/c,\) as a datapoint may thus be described by a graph \(G_{y_i}^c = (V_{y_i}^c, E_{y_i}^c)\). 
To uniformize the graph size of each datapoint  $y_i$, and robustify its representation, we highlight the most relevant features by  \(L\) self-attention heads \(SA.\text{Head}_j\) to glean feature maps \(\{f^c_{i,j}\}\), where \(c\) denotes class, \(i\) datapoint index (\(i \in \{1, \ldots, n/c\}\)), and \(j\) feature index (\(j \in \{1, \ldots, L\}\)). Each size-standardized graph \(G_{y_i}^c\in \mathcal{G}^c\) (set of all graphs for class $c$) is trained to maximize its Laplacian \(\mathcal{L}\) spectral gap \(\lambda_2 - \lambda_1\) to yield a well connected  representative class code-graph \(\hat{G}^c\)\cite{dey2023dlkoopman}.\footnote{For \(G = (V, E)\) with \(|V| = L\), \(\mathcal{L} = D - A\), where \(D_{ii}\) is vertex \(i\)'s degree, and \(A\) is the adjacency matrix. \(\mathcal{L}\)'s eigenvalues are \(\lambda_1, \lambda_2, \ldots, \lambda_L\), with \(\lambda_1 = 0\) for connected graphs.} Self-attention calculates feature importance, capturing dependencies in \(y_i\):

\begin{minipage}{8.5cm}
\begin{equation}
\small
\begin{aligned}
f^c_{i,j} = SA&.\text{Head}_j(y_i),\\
&\forall j \in \{1,\ldots,L\},
\forall i \in \{1,\ldots,n/c\},
\forall c \in \{1,\ldots,C\}.
\end{aligned}
\end{equation}
\end{minipage}

This step corresponds to ${\mathcal A}: {\mathbb R}^{n\times d} \rightarrow {\mathbb R}^{n\times d\times L}$. To avoid including nuisance and non-characteristic background information  \(f^c{i,j}\), we proceed to first  crop: \( f^{c'}{i,j} = \text{Att. Crop}(f^c{i,j}) \) prior to transferring them into \(Y'\) using \(\text{ConvT}\) (Conversion Transformation), guided by \(\hat{G}^c = (V_{\hat{G}^c}, E_{\hat{G}^c})\). The dimension expansion  constructively generates a new datapoint \(y'{\text{new}}\) by selecting one component from each column of the matrix of features $F=[f{ij}^{c'}] \in {\mathbb R}^{n/c\times L}$ as vertices of ${\hat{G}^c}$. More generally, we can express this as ${\hat{G}^c}=\left( V_{new}^c, E_{\hat{G}^c}\right)$ where $V_{new}^c = S \odot F$. Here, $S \in {\mathbb R}^{n/c \times L}$ is a selection matrix with exactly one 1 in each column and zeros elsewhere, and $\odot$ denotes the Hadamard (element-wise) product. To introduce variability in the selection process, we can define $S = P^i M$, where $P^i \in {\mathbb R}^{n/c \times n/c}$ is a permutation matrix that shuffles the rows, and $M \in {\mathbb R}^{n/c \times L}$ is a fixed matrix with exactly one 1 in each column (in different rows) and zeros elsewhere. The index $i$ in $P^i$ allows for different permutations in each iteration for new datapoint generation. Eq. 4 integrates selected features with the code-graph, ensuring accurate feature placement:

\begin{minipage}{8.5cm}
\begin{equation}
    \begin{aligned}
    y'_{\text{new}} &= \sum_{j=1}^L \text{ConvT}(f^{c'}_{i,j}, \hat{G}^c),
    \end{aligned}
\end{equation}
\end{minipage}

This step corresponds to ${\mathcal E_G}: {\mathbb R}^{n\times d\times L} \rightarrow {\mathbb R}^{n'\times d}$. The "ConvT" implements the mapping the enhanced features \(f^{c'}_{i,j}\) to nodes \(V_{\hat{G}^c}\) of Expansive Graphs \ of \(\hat{G}^c\), while ensuring \(v' \in V_{y'_i}\) inherit \(G_{y_i}^c\) good connectivity properties \cite{dey2023dlkoopman, villani2009optimal}. \(\hat{G}^c = (V_{\hat{G}^c}, E_{\hat{G}^c})\) maintains robust connectivity with minimal edges. Embedding \(f^{c'}_{i,j}\) through \(\hat{G}^c\), preserves \(\lambda_{{\hat G}_{{y}_{i,k}}} \approx \lambda_{G_{{y'}_{i,k}}}\), while maintaining structural and training-valued robustness properties.
One additionally notes that $L$ as the number of features used for generating a new datapoint, from  the minimal $n$ samples, results  in an Expander Graph Mapping yielding $n^L - n$ unique combinations.  $n^L$ represents the total possible combinations excluding  $n$ original samples in the dataset.
As we elaborate next, robust consistency and diversity of the generative data \(Y'_{\text{new}}\) relative to the original small dataset are computationally preserved by regularization of the overall loss.$\qedsymbol$  

\subsection{Loss Objective Functions and Regularization}
The generative ES flow in Fig.~\ref{fig:1}(b) is sought by regularizing   global losses with other local constraints. Global losses include data reconstruction and classification, while local  regularizing constraints account for distributional consistency (Wasserstain distance), diversity, and feature refinement.

\subsubsection{Total Loss (\(\mathcal{L}_{\text{total}}\)):} 

{\bf Claim:} {\bf \it The optimal expansion of a dataset \(X\) to \(X'\) can be achieved by minimizing a total loss function \(\mathcal{L}_{\text{total}}\), composed of four terms working in concert:}
\vspace{-5pt}
\begin{equation}
\vspace{-5pt}
   \mathcal{L}_{\text{total}} = \alpha_0 \mathcal{L}_{re} + \alpha_1 \mathcal{L}_{ce} + \alpha_2 \mathcal{L}_{\mathcal{W}} + \alpha_3 \mathcal{L}_{cov}
\end{equation}
where \(\mathcal{L}_{re}\) is the reconstruction loss ensuring accuracy of the expanded dataset, \(\mathcal{L}_{ce}\) is the classification loss maintaining label consistency, \(\mathcal{L}_{\mathcal{W}}\) is the Wasserstein distance preserving distributional consistency, \( \mathcal{L}_{cov} \) is the covariance loss promoting diversity in the expanded dataset, and \( \alpha_0, \alpha_1, \alpha_2, \) and \( \alpha_3 \) are balancing hyperparameters. These terms jointly ensure that the synthesized samples are diverse, persistent, and maintain a good cover of the original data distribution while preserving essential discriminative features.

\subsubsection{Reconstruction Loss (\(\mathcal{L}_{re}\)):} The AE parametrization denoted by \(\phi\) and \( \phi^{-1} \) for encoder and decoder respectively, follows the standard optimization loss between the input and output distribution written as:

\begin{equation}
\small
\begin{aligned}
\mathcal{L}_{re}(\phi, \theta ; X)= 
&\mathbb{E}_{Y \sim q_\phi(Y \mid X)}[-\log p_\theta(X \mid Y)] \\
&+ \text{KL}[q_\phi(Y \mid X) \Vert p(Y)], 
\end{aligned}
\end{equation}

where the first term is the expected negative log-likelihood, and the second term is the KL divergence between the encoded distribution \(q_\phi(Y | X)\) and a prior distribution \(p(Y)\) \cite{goodfellow2014generative}. To selectively exploit the most relevant features and further refine the intrinsic linear evolution of the nonlinear dynamics, we induce a composition map of a multi-head attention transformation on \(Y\) followed by  an expander graph mapping to yield \(Y' \in \mathbb{R}^{n' \times d}\) \cite{vaswani2017attention}. The expansion phase entails crafting  \( Y' \) assume a  higher dimension (\( n' \)) while reflecting the original dataset distribution. The latter  is guided by the minimization of the Wasserstein distance \citep{vaserstein1969markov} \( \mathcal{W}(Y, Y') \), ensuring the safeguard of the distributional integrity of \( Y \).

\subsubsection{Distributional Consistency(\(\mathcal{L}_{\mathcal{W}}\)):} Wasserstein distance, a key expression in  Optimal Transport Loss \cite{villani2009optimal}, measures the cost to align the distribution of encoded data \(Y\) with expanded representation \(Y'\). It quantifies the minimal effort to morph \( p_Y \) into \( p_{Y'} \), measuring dataset expansion effectiveness \cite{roheda2023fast}. The Wasserstein distance \citep{vaserstein1969markov} is:
\begin{minipage}{8.5cm}
\begin{equation}
    \begin{aligned}
    \mathcal{L}_{\mathcal{W}}(p_{Y}, p_{Y'}) = \min_{\pi \in \Pi(p_{Y}, p_{Y'})} \iint c(Y, Y') \pi(p_{Y}, p_{Y'}) \, dY \, dY'
    \end{aligned}
\end{equation}
\end{minipage}
In our formulation, \( \pi \) is the OT plan associating \( p(Y) \) and \( p(Y') \). Minimizing \(\mathcal{L}_{\mathcal{W}}\) ensures \( Y' \) statistically resemble \( Y \) and preserve its geometric and topological properties, crucial to maintaining data distributional fidelity for subsequent learning tasks dependent on the data' manifold structure \cite{villani2009optimal}. \( Y' \) is mapped back to a high-dimensional image space using \( \phi^{-1} \), resulting in \( X' \in \mathbb{R}^{n' \times D} \). Note that using the same AE for encoding and decoding ensures \( X' \) be a plausible AE output, retaining the original dataset's structure and distributional properties. A classifier \( Cl \)-trained on \( X' \) predicts labels \( \hat{l} \) as if trained on \( X \), benefiting from \( X' \)'s distilled information and enabling efficient training with reduced data.

\subsubsection{Classification Loss (\(\mathcal{L}_{ce}\)):} The classification loss provides an implicit feedback, measuring discrepancy between predicted and true labels to maintain label consistency. It uses both original (\(X\)) and synthesized (\(X'\)) images, merged and passed through the classifier. The Cross-Entropy (CE) loss is used:
\begin{minipage}{8.5cm}
\begin{equation}
    \mathcal{L}_{ce}(Cl, \tilde{X}, l) = -\sum_{i} l_i \log(Cl(\tilde{X}_i))
\end{equation}
\end{minipage}
Here, \(Cl\) is the classifier, \(\tilde{X}\) the combined set of original and reconstructed data, \(l\) the vector of true labels, and \(\tilde{X}_i\) the \(i\)-th data instance in the merged dataset \cite{goodfellow2014generative}. This loss component ensures the expanded dataset encapsulate the original data's structural attributes and label characteristics, while preserving essential discriminative features and preventing categorical information dilution in the course of expansion.
It enables scalable training on large datasets while maintaining performance comparable to full-dataset training.

\subsubsection{Expansion Diversity Loss (\( \mathcal{L}_{cov} \)):} Including covariance constraint term aims at promoting diverse and representative expanded datasets. This regularizer encourages distinct features in latent representations \( Y \) to enhance spread and space explorating persistence of generated data representation. 
Mathematical definition of Cov. Loss is:

\begin{minipage}{8.5cm}
\begin{equation}
    \mathcal{L}_{cov}(Y') = \left\| \text{Cov}(Y') - I \right\|_F^2,
\end{equation}
\end{minipage}

where \( \text{Cov}(Y') \) is the covariance matrix of \( Y' \), \( I \) is the identity matrix, and \( \left\| \cdot \right\|_F \) is the Frobenius norm.$\qedsymbol$

\vspace{-10pt}
\noindent
\begin{minipage}[t]{0.48\textwidth}
\scriptsize 
\begin{algorithm}[H]
\fontsize{7.5}{10}\selectfont 
\caption{Expansive Syn. Train Algorithm}
\begin{algorithmic}[1]
\State \textbf{Given:}
\State $X \in \mathbb{R}^{n \times D}$, original dataset with $n$ samples
\State AE: $\phi$ (Encoder mapping $\mathbb{R}^D \to \mathbb{R}^d$) and $\phi^{-1}$ (Decoder mapping $\mathbb{R}^d \to \mathbb{R}^D$)
\State MHSSA: Multi-head Spatial Self-attention
\State EGM: Expander Graph Mapping
\State $\alpha_0, \alpha_1, \alpha_2, \alpha_3$: Weights for loss components
\State $\text{N}$: Number of training epochs
\State $\text{M}$: Number of classes of data
\State \textbf{Initialize:} Pretrained Parameters of AE ($\phi$, $\phi^{-1}$), Classifier $Cl$
\For{$\text{epoch} = 1$ \textbf{to} $\text{N}$}
    \For{$\text{class} = 1$ \textbf{to} $\text{M}$}
        \State $Y \gets \phi(X)$
        \State $Y_{SA} \gets \text{MHSSA}(Y)$
        \State $Y' \gets \text{EGM}(Y_{SA})$
        \State $X' \gets \phi^{-1}(Y')$
        \State $L_{re} \gets ||X' - X||^2$
        \State $\hat{Y} \gets Cl(X \oplus X'), (\oplus \text{: concatenation})$
        \State $L_{ce} \gets -\sum_i l_{i} \cdot \log(\hat{Y}), (l_{i} \text{: true labels})$
        \State $\mathcal{L}_{\mathcal{W}} \gets \mathcal{W}(Y, Y'), (\mathcal{W} \text{: Wasserstein Distance})$
        \State $\mathcal{L}_{cov}(Y') \gets \left\| \text{Cov}(Y') - I \right\|_{F}^2$
        \State $L_{total} \gets \alpha_{0} L_{re} + \alpha_{1} L_{ce} + \alpha_{2} \mathcal{L}_{\mathcal{W}} + \alpha_{3} \mathcal{L}_{cov}$
        \State \text{Update Parameters}
    \EndFor
\EndFor
\end{algorithmic}
\end{algorithm}
\end{minipage}\hfill

\section{Experiments and Results}

\subsection{Stages of Implementation}
\begin{figure}
  \centering
  \includegraphics[width=9cm]{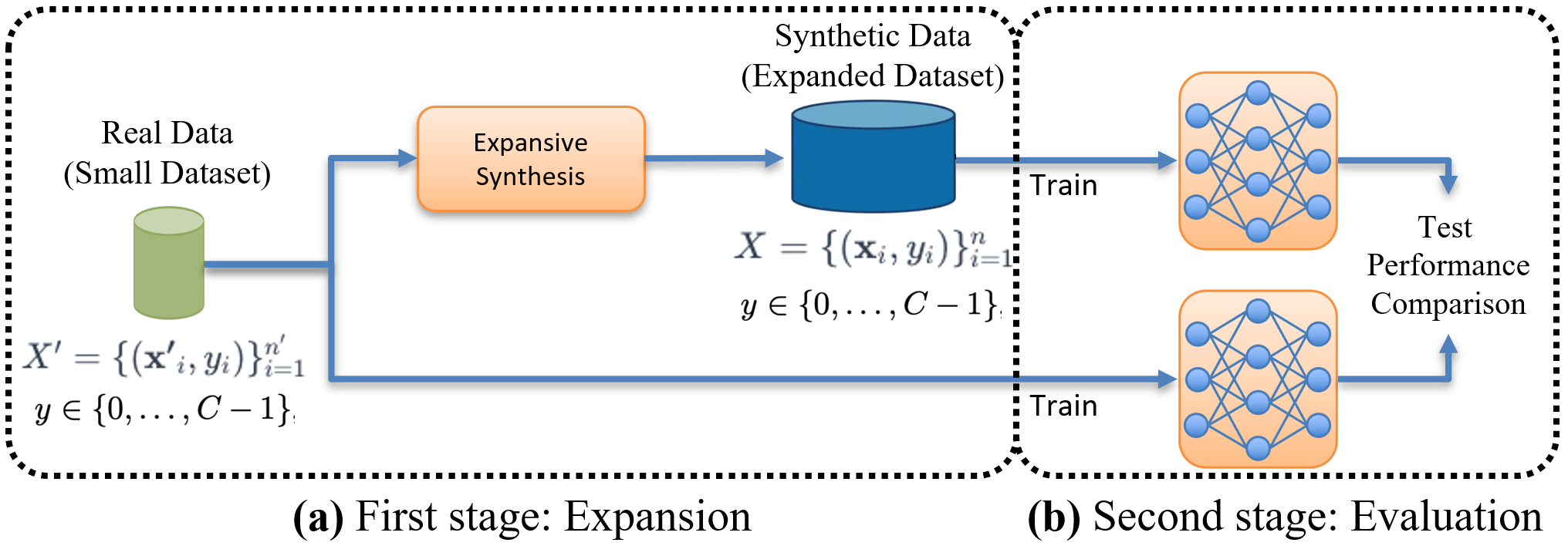}
  \vspace{-20pt}
  \caption{Stages of Implementation and evaluation of a expansion model}
  \vspace{-15pt}
  \label{fig:3}
\end{figure}

Fig.~\ref{fig:3} illustrates a two-phase process: expansion and evaluation.

\subsubsection{First Stage (Expansion):} We start with a small dataset \( X \) of \( n \) pairs \( (x_i, l_i) \), where \( x_i \) is input data and \( l_i \) are labels ranging from 10 to \( c \). \( X \) is expanded to create \( X' \), a larger synthesized dataset of \( n' \) pairs \( (x'_i, l_i) \), where \( x'_i \) are synthesized datapoints (expanded version) and \( l_i \) are labels, maintaining the original label range.

\subsubsection{Second Stage (Evaluation):} The synthesized dataset \( X' \) is used to train a classifier. We train the same type of classifier on the corresponding small dataset \( X \).
The performance of both classifiers is evaluated on a test set, and as we show, the goal of training for inference  on \( X' \) outperforms that carried out on  \( X \) and its augmented version (using current techniques) is achieved.   
The hypothesis is that if \( X' \) is a good expansion of \( X \), the \( X' \)-trained classifier should generalize almost as well as the \( X \)-trained one on unseen data, demonstrating that \( X' \) captures \( X \)'s core information, enabling effective training with less data.

\subsection{Results and Comparisons}
\begin{table}[ht]
  \caption{Classification accuracy results (\%) initiated with 10 and 100 images per class (Img/Cls) post expansion for our expansive synthesis model with different settings on different datasets. Acronyms: RwA (Row-wise Attention), SA (Self-Attention), SC (Skip Connections), MHSA (Multi-Head Self-Attention), Lin. Int. (Linear Interpolation), EGM (Expander Graph Mapping).}
  \centering
  \includegraphics[width=9cm]{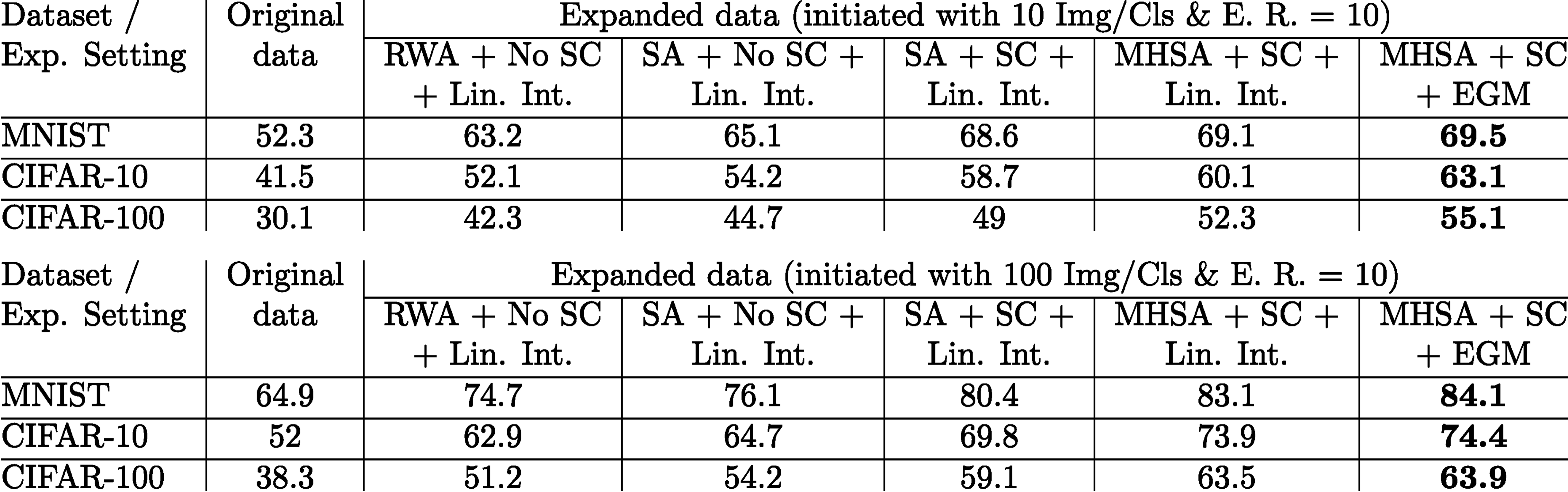}
  \vspace{-20pt}
  \label{fig:4}
\end{table}

The effectiveness in generating large-scale, high-performing datasets from minimal samples is demonstrated by comparing classifiers trained on these synthesized datasets against those trained on original and also on classically augmented datasets. Tables~\ref{fig:4} compare classification accuracies across MNIST \citep{deng2012mnist}, CIFAR-10, and CIFAR-100 \citep{krizhevsky2009learning}. Classifiers trained on 10 original images achieve the lowest accuracies. Expansive synthesis with row-wise self-attention improves accuracy by an average 20\% over the baseline. Spatial attention further enhances performance by 6\%, while adding skip connections yields an additional 8\% improvement. Multi-head spatial self-attention boosts results by 4\%, and incorporating Expander Graph Mapping (EGM) achieves the best results with a 6\% enhancement over the previous configuration. The ablation study demonstrates that each module incrementally improves performance, with multi-head self-attention and EGM optimizing feature representation.
Table~\ref{fig:6} compares classification accuracy for models initialized with 10 or 100 images per class across MNIST, CIFAR-10, and CIFAR-100. The expansive synthesis model consistently outperforms traditional augmentation. On MNIST, accuracy improves from 52.3\% to 54.7\% with 10 images per class and from 64.9\% to 66.8\% with 100 images per class. Similar improvements are seen for CIFAR-10 and CIFAR-100. These results demonstrate our model's effectiveness in enhancing dataset quality and classifier performance.
\vspace{-10pt}
\begin{table}[ht]
  \caption{Comparison of classification acc. (\%) for models initiated with 10 and 100 images per class (Img/Cls) across different datasets. The table compares the performance of classifiers trained on original data, classically augmented data, and data expanded using our ES method, all with the same Expansion Ratio (ER=10).}
  \centering
  \includegraphics[width=9cm]{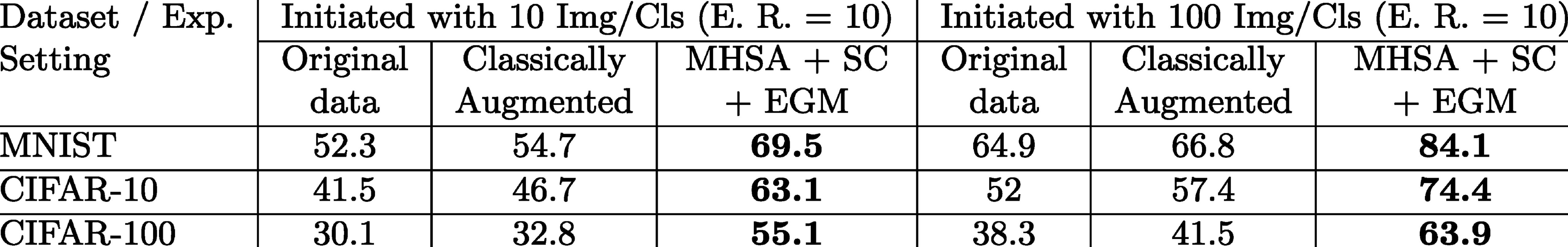}
  \vspace{-20pt}
  \label{fig:6}
\end{table}
\section{Conclusion and Future Works}
In this paper, we introduced the Expansive Synthesis model for generating large-scale, high-performing datasets from minimal samples. It uses a Koopman feature space within an AE to transform non-linear latent space features into a linear space. Composing a multi-head spatial self-attention mechanism for improving feature extraction, with  Expander graph mappings to secure feature connectivity during expansion, together with proper regularization such as Optimal transport preserving distribution integrity, and other, yields a principled, effective enrichment of small datasets destined for training to learn. This, recall, is an effective alternative solution to augmentation and synthetic data complementation/replacement which has been shown and even GANs which demand large amounts of data for training and still susceptible to instability. 

\newpage
\bibliographystyle{plainnat}
\bibliography{ICASSP_2025.bib}

\end{document}